\documentclass[10pt,twocolumn,letterpaper]{article}

\usepackage{wacv}              

\usepackage{graphicx}
\usepackage{amsmath}
\usepackage{amssymb}
\usepackage{booktabs}

\usepackage[table]{xcolor}
\usepackage[accsupp]{axessibility} 
\usepackage[pagebackref,breaklinks,colorlinks]{hyperref}

\usepackage[capitalize]{cleveref}
\crefname{section}{Sec.}{Secs.}
\Crefname{section}{Section}{Sections}
\Crefname{table}{Table}{Tables}
\crefname{table}{Tab.}{Tabs.}

\def\wacvPaperID{2376} 
\def\confName{WACV}
\def\confYear{2024}

\begin{document}

\title{Sharp-NeRF: Grid-based Fast Deblurring Neural Radiance Fields Using Sharpness Prior}

\newcommand\CoAuthorMark{\footnotemark[\arabic{footnote}]}
\newcommand\CorrespondingAuthorMark{\footnotemark[\arabic{footnote}]}
\author{
    Byeonghyeon Lee$^1$\thanks{Equal contributions} \and
    Howoong Lee$^{2,3}\protect\CoAuthorMark$ \and
    Usman Ali$^{2}$\thanks{Corresponding authors} \and
    Eunbyung Park$^{1,2}\protect\CorrespondingAuthorMark$ \and
    $^1$Department of Artificial Intelligence, Sungkyunkwan University \\ 
    $^2$Department of Electrical and Computer Engineering, Sungkyunkwan University \\
    $^3$Hanwha Vision \\
}

\maketitle

\begin{abstract}
Neural Radiance Fields (NeRF) has shown its remarkable performance in neural rendering-based novel view synthesis. However, NeRF suffers from severe visual quality degradation when the input images have been captured under imperfect conditions, such as poor illumination, defocus blurring and lens aberrations. Especially, defocus blur is quite common in the images when they are normally captured using cameras. Although few recent studies have proposed to render sharp images of considerably high-quality, yet they still face many key challenges. In particular, those methods have employed a Multi-Layer Perceptron (MLP) based NeRF which requires tremendous computational time. To overcome these shortcomings, this paper proposes a novel technique Sharp-NeRF---a grid-based NeRF that renders clean and sharp images from the input blurry images within a half an hour training. To do so, we used several grid-based kernels to accurately model the sharpness/blurriness of the scene. The sharpness level of the pixels is computed to learn the spatially varying blur kernels. We have conducted experiments on the benchmarks consisting of blurry images and have evaluated full-reference and non-reference metrics. The qualitative and quantitative results have revealed that our approach renders the sharp novel views with vivid colors and fine details, and it has considerably faster training time than the previous works. Our code is available at \href{https://github.com/benhenryL/SharpNeRF}{https://github.com/benhenryL/SharpNeRF}.
\end{abstract}

\begin{figure}[t]
\begin{center}
\includegraphics[width=1.0\linewidth]{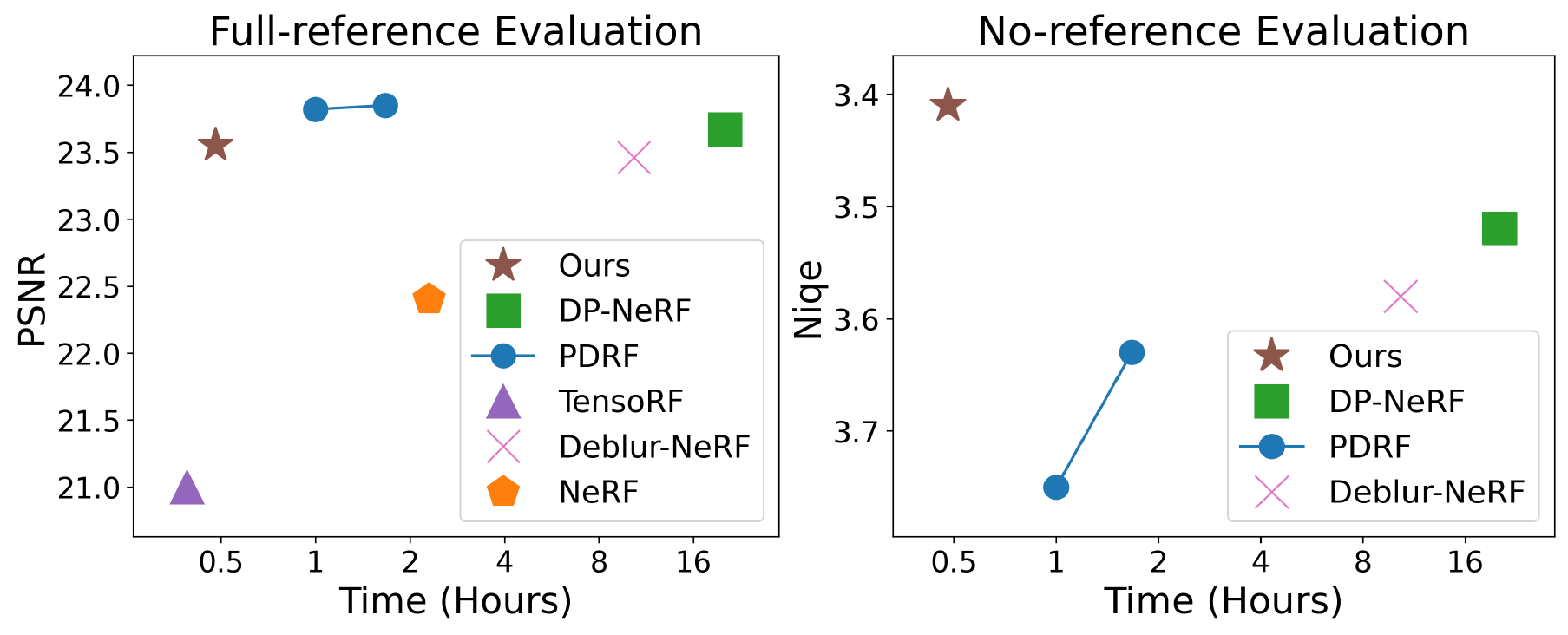}
\end{center}
\vspace{-0.5cm}
   \caption{Comparison in terms of training time and image quality on the real defocus dataset. Left: Evaluated under full-reference metric (PSNR). Right: Evaluated under no-reference metric (Niqe). }
\label{fig:curve}
\end{figure}
\section{Introduction}

In the field of computer vision and graphics, with numerous applications in visual effects, e-commerce, AR/VR, and robotics, it is essential to model and reconstruct 3D scenes as representations that can synthesize high-quality images. Recently, Neural Radiance Fields (NeRF)~\cite{nerf} has been proposed, and it has shown excellent performance in 3D scene representation and rendering photorealistic images for novel views. While successful, many existing NeRF methods assume that the training images are obtained in perfect condition, which bypasses vital issues that would arise in many practical settings. For example, training images captured with commodity hand-held devices inevitably contain a considerable amount of blur, e.g., motion or defocus blur~\cite{liu2020estimating,suwajanakorn2015depth}. This violates 3D view consistency between different view images during training and deteriorates the rendered image quality.

A few approaches have been recently suggested to address the issue~\cite{ma2022deblur, lee2022deblurred, peng2022pdrf}. For example, Deblur-NeRF~\cite{ma2022deblur} proposed a deblurring neural radiance fields that renders clean images at novel views from blurry input images. It proposed to generate clean images first and apply convolution operation with the learnable blur kernels. At the time of inference, blur kernels are not utilized, and hence, it enables NeRF to produce sharp images. DP-NeRF~\cite{lee2022deblurred} improved Deblur-NeRF by imposing 3D consistency to blur kernels and using depth information. Although promising results have been shown, they suffer from very long training and inference times due to the large number of neural network propagations to render images, please see Fig.~\ref{fig:curve}.

In this work, we suggest Sharp-NeRF, a fast NeRF-based framework for rendering sharp images with blurred training images. Inspired by the recent approaches that incorporate additional data structures, such as voxel grids~\cite{chen2022tensorf, kplanes_2023, fridovich2022plenoxels, Hexplane}, trees~\cite{yu2021plenoctrees, realtime_plenoctree} and hashes~\cite{muller2022instant, f2nerf}, we leverage a decomposed-grid representation for neural fields architecture~\cite{chen2022tensorf} to accelerate the training time and efficiently represent the colors and densities.

In addition, we propose \textit{discrete learnable kernels} (without neural networks) for blurring convolution operations as opposed to the existing methods that utilize an extra MLP to generate the blur kernels. Using a neural network as a kernel-generating function introduces additional computational costs, and the bias induced by the neural network may prevent it from learning complex and sophisticated kernels that are necessary to handle various real-world scenarios. Furthermore, it has been known that the size of neural networks, e.g., width or depth, determines the complexity of the functions they can represent. Therefore, more optimal kernels require larger neural networks, which results in adding more computational burden during training.

\begin{figure*}[t]
\begin{center}
\includegraphics[width=1.0\linewidth]{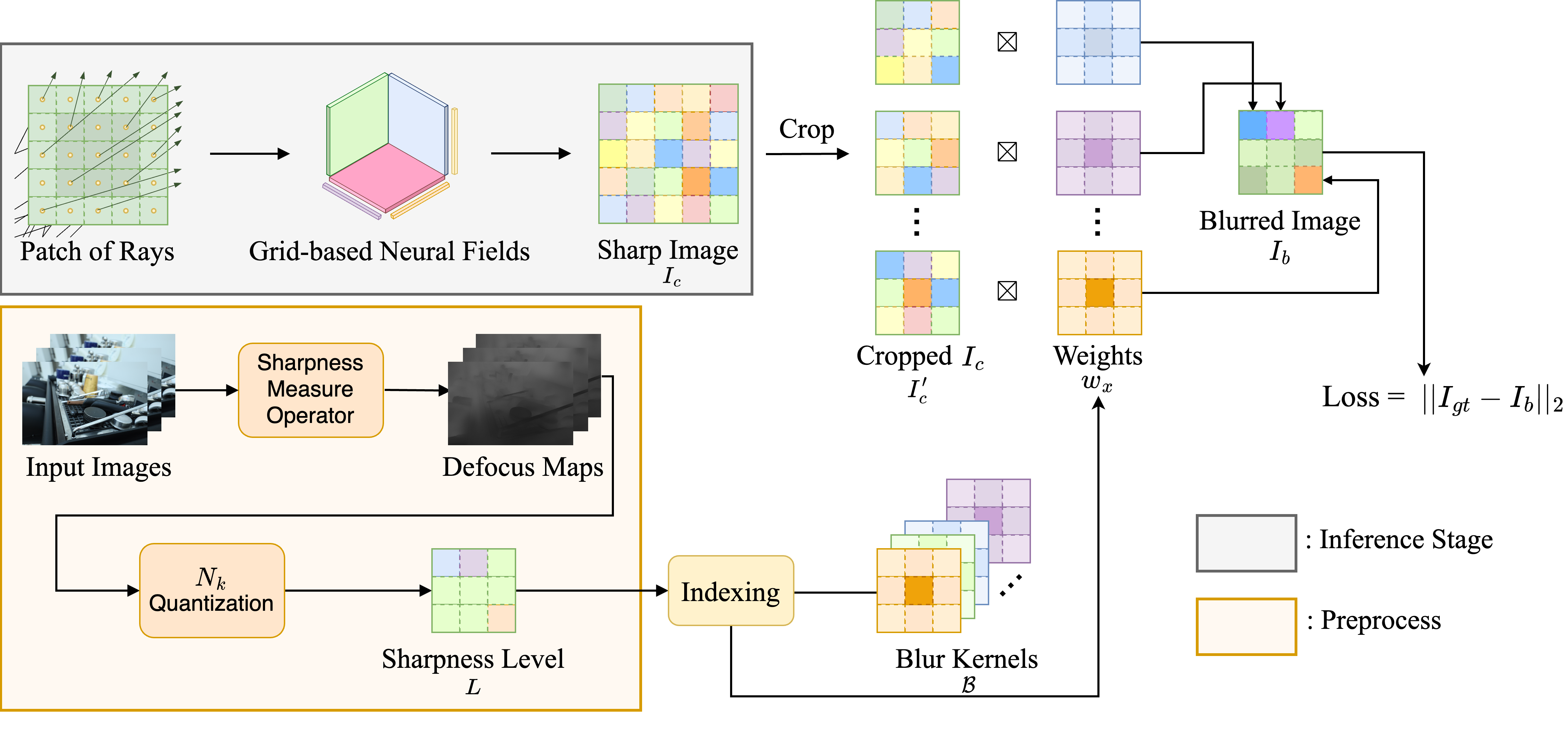}
\end{center}
\vspace{-0.5cm}
   \caption{The overall architecture of Sharp-NeRF. $\boxtimes$ stands for weighted sum. First, it computes defocus map of each training view using sharpness measure operator. Then the defocus map is quantized into $N_k$ values which is used as per-pixel sharpness level map $L$. This is a preprocess and is not required to compute $L$ druing process. During training, it takes ray patches as inputs to backbone neural fields model~\cite{chen2022tensorf} and render sharp and clean image $I_c$. $I_c$ is then cropped into several small patches $I_C'$ with stride 1 where size of each patch is $K \times K$. Preprocessed $L$ of the input ray patch is also given as input data and used to indexing blur kernels $\mathcal{B}$ to obtain per-pixel weight $w_{x}$. Subsequently, $w_{x}$ and $I'_c$ are weighted sum to render blurred image $I_b$. Note that at the time of inference, only modules in gray box are valid which means blur kernel is no longer required and only $I_c$ is used as a final rendered outcome.   }
\label{fig:architecture}
\end{figure*}

As blurriness varies spatially across the scene, the ideal approach would involve retaining blur kernels for every individual pixel in every image. However, this becomes impractical due to the current prevalence of high-resolution images and multiple-view images. We introduce a sharpness prior to the proposed framework that enables us to maintain only a few number of discrete kernels. We propose to measure the per-pixel sharpness level and assign a blur kernel to a group of pixels whose sharpness level is similar. This strategy ensures that the blur kernels learn the spatially varying sharpness consistently and reliably with a marginal increase in learnable parameters. The overall architecture of our proposed method is shown in Fig.~\ref{fig:architecture}. 

To optimize the training time further, we propose random patch sampling during training. As the blurriness is attributed to the intermingling of neighboring pixels, the existing deblurring methods render multiple neighboring pixels (usually 5 or 6) to render one pixel. It increases the training time proportionally to the number of rays per pixel. We develop a training strategy that renders a patch instead of a pixel and apply the convolution operation over the patch with a learned discrete kernel to produce the blurred patch. This strategy significantly reduces the training time while maintaining the rendered image quality.

We have tested the proposed approach on a challenging real-world dataset consisting of blurry training images. The experimental results have shown that Sharp-NeRF improves the training speed by a large margin and performs on par with the existing deblurring methods regarding visual quality. We also demonstrated that the proposed method outperforms the prior arts in more representative quantitative metrics, such as Brisque and Niqe, widely used in measuring visual quality.

To sum up, our contributions are the following:
\begin{itemize}
    \item We proposed novel learnable grid-based kernels to obtain sharp output from neural radiance fields. These kernels are optimized directly without requiring additional networks.
    \item For better and more reliable kernel optimization, a sharpness prior has been incorporated that explicitly measures the sharpness of pixels.
    \item We proposed to adopt random patch sampling to reduce the computational complexity of rendering.
    \item Novel view images rendered by our method are sharp, contain vivid colors and fine details, and have been achieved through the training time of a half an hour.
\end{itemize}


\section{Related Works}
\subsection{Neural Radiance Fields}     
Recently, the Neural Radiance Fields (NeRF) has emerged as a powerful approach for synthesizing high-fidelity 3D scenes from 2D images. NeRF leverages deep neural networks to represent volumetric scene properties and enables realistic rendering from novel viewpoints. 
A radiance fields is a continuous function $f$ that maps a 3D location $x \in \mathbb{R}^3$ and a viewing direction $d \in \mathbb{S}^2$ to estimate density $\sigma \in [0,\infty)$ and a color value $c \in [0, 1]^3$ of the given point.  A multi-layer perceptron (MLP) has been used by \cite{nerf} to parameterize this function, where the weights of MLP are optimized to reconstruct a set of input images of a certain scene:
\begin{equation}
    f_\theta : \big(\gamma(x), \gamma(d) \big) \rightarrow (c, \sigma)  \; .
\end{equation}
Here, $\theta$ denotes the network weights, and $\gamma$ is a predefined positional encoding~\cite{tancik2020fourier} applied to $x$ and $d$. Given the volume density and color of points, volume rendering~\cite{volume_rendering} is used to produce the images at novel views~\cite{nerf}. 

With the success of NeRF, many follow-up studies have been conducted that improve the one or multiple modules of NeRF for improved quality of rendering. 
Several approaches, e.g., grid-based F2-NeRF~\cite{f2nerf} and TensoRF~\cite{chen2022tensorf}, aim to improve the rendering speed~\cite{reiser2021kilonerf, hu2022efficientnerf, fridovich2022plenoxels,yu2021plenoctrees,muller2022instant,sun2022direct,hedman2021snerg,garbin2021fastnerf}. Few others, e.g., FreeNeRF~\cite{Yang2023FreeNeRF}, PixelNeRF~\cite{yu2021pixelnerf} and MVSNeRF~\cite{chen2021mvsnerf}, try to predict a continuous neural scene representation conditioned on one or a few input images and SNeRG~\cite{hedman2021snerg}, MeRF~\cite{Reiser2023MERF}, and BakedSDF~\cite{yariv2023bakedsdf} bake model to achieve real-time view synthesis. 

\noindent\textbf{Learning-based deblurring networks for NeRF.}  
Some studies in the literature have explored a learning-based approach to directly train the neural networks to deblur the NeRF-based rendered images. These methods leverage a pair of blurred and clear image datasets to learn the mapping between the blurred rendering and the corresponding sharp image. 
However, many of these tasks use MLP-based architectures. The Deblur-NeRF~\cite{ma2022deblur} and DP-NeRF~\cite{lee2022deblurred} take a long time to train a MLP-based neural fields and kernel. The Hybrid Neural Rendering Model~\cite{dai2023hybrid} has a grid kernel, but it is fixed and applies only to the camera motion blur~\cite{wang2022bad}. In the defocus blurred scene, point spread function (PSF) is spatially varying and cannot be applied straightforward. 
DoF-NeRF~\cite{dof-nerf} has a limitation that all-in-focus images are required for training. Although PDRF~\cite{peng2022pdrf} is similar to our model in that it used grid-based representation, it still uses two MLPs; first to predict the scene density and the second to build blur kernels. In contrast, we propose a well-curated approach Sharp-NeRF---a grid-based model that can render sharp novel images from blurry input images within half an hour of training.


\subsection{Image Blind Deblurring}           
It is common to observe that some parts of the pictures are blurred when we normally capture them using optical imaging devices. Among a number of factors that cause this blurriness are lens defocusing, camera shake, and fast motion~\cite{ruan2022learning,abuolaim2022improving}.
The out-of-focus or defocus blur occurs in the imaging process when the image plane is away from the ideal reference plane. 
Mathematically, the deterioration of an image due to defocus blur is typically represented by $B = I \otimes K  + \epsilon  \, ,$
where $I$ is the latent sharp image, $B$ is the observed blurry image, $K$ is the blur kernel, $\otimes$ is the convolution operator, and $\epsilon$ is the additive white Gaussian noise that often appears in natural images. The earliest canonical method that deconvolves an image is \cite{richardson1972bayesian} where a known PSF---the blur kernel---is used to iteratively minimize an energy function to get a maximum likelihood approximation of the original image. Usually, both the sharp image $I$ and blurring kernel $K$ are unknown. Recovering the latent sharp image given only a single blurry image $B$ is known as image blind deblurring~\cite{whyte2012non}. This is a highly ill-posed and long standing problem in the image and vision community. In this process, first a blurring kernel is estimated which is then used to derive the deblurred image.
A variety of methods have been proposed in the literature to tackle the blurring problem. The traditional approaches rely on natural image priors and often formulate deblurring as an optimization problem \cite{xu2013unnatural,pan2016blind,liu2020estimating,zhang2022pixel}. On the other hand, most of the deep learning based methods directly map the blurry image with the latent sharp image by employing convolutional neural networks (CNN) \cite{nimisha2017blur, zhang2019deep, ren2020neural}. However, most of the learning-based approaches disregard the underlying imaging process, and hence they perform less effectively in the case of severe blur. 


\subsection{Sharpness Measurement}\label{sec:sharpness_measurement}     
Computing the sharpness level of image pixels is a crucial step for many imaging applications, including the object detection and shape from focus (SFF) system \cite{ali2021robust,yang2022deep,won2022learning}. In the case of partially out-of-focus image, focused areas typically exhibit greater intensity variations than defocused areas. Any operator can be used as a sharpness measure as long as it can distinguish between sharp (focused) and blurred (defocused) pixels in a partially blurred (defocused) image. The sharpness measurements or levels of image pixels are obtained by convolving the image with the sharpness measure. A detailed analysis of sharpness measures for focus computation has been provided in \cite{pertuz2013analysis, gallego2019focus}. Among different categories of sharpness measures, derivative-based measures appear to be an interesting option because of their inherent ability to capture the rate of change.
A well-known and commonly used focus measure operator is modified laplacian (ML)~\cite{nayar1994shape}.
Inspired by the effectiveness of sharpness measures, we propose to leverage a learning-based sharpness prior in our deblurring approach for NeRF. Our analysis has shown that enforcing the sharpness prior is effective in rendering better deblurred images.

\section{Method}
\subsection{Preliminary: Tensorial Radiance Fields}        
We adopted Tensorial Radiance Fields (TensoRF)~\cite{chen2022tensorf} as our backbone grid-based neural radiance fields. It encodes and reconstructs the scene using two 3D grids $G_{\sigma}$ and $G_c$ that estimate density and color respectively, which are further decomposed to several low-rank 2D and 1D tensors as following:
\begin{equation} \label{eq:TensoRF_grid}
    G_{\sigma} = \sum_{i=0}^{N_x}v_i^x \circ M_i^{y,z} + \sum_{i=0}^{N_y}v_i^y \circ M_i^{x,z} + \sum_{i=0}^{N_z}v_i^z \circ M_i^{x,y},
\end{equation}
where $N_x$ is the number of the decomposed tensors, $v^x$ is 1D vector along the x-axis, $M^{y,z}$ is 2D matrix along the y- and z-axis. The grid for color estimation $G_c$ is defined identically to $G_{\sigma}$. TensoRF takes input coordinates $x \in \mathbb{R}^3$ and viewing directions $d \in \mathbb{S}^2$, and produces 4D vectors comprising of density and RGB values, such as,
\begin{align} \label{eq:TensoRF_grid}
    &\sigma(r) = \zeta(G_{\sigma}(x)), \\
    &c(r) = f_\theta(G_c(x), d),
\end{align}
where $\theta$ denotes the parameters of shallow MLP for color rendering and $\zeta$ denotes softplus activation function~\cite{dugas2000incorporating} .
Then these $\sigma(r)$ and $C(r)$ are finally used to compute pixel intensity using volume rendering~\cite{volume_rendering}:
\begin{align} \label{eq:TensoRF_grid}
    &C(r) = \sum_{q=1}^Q T_q(1 - \exp(-\sigma(r)_q\Delta_q))c(r)_q, \\
    &T_q = \exp(-\sum_{p=1}^{q-1}\sigma(r)_p\Delta_p),
\end{align}
where $C(r)$ is the approximated intensity of ray $r$, $q$ is sampled points, and $\Delta_p$ is the distance between pixels.
\subsection{Patch Sampling}     
Random ray sampling has been widely used in NeRF literature~\cite{nerf,lin2021barf,wang2021nerf}. This approach selects rays arbitrarily across the scene, where it is highly unlikely that it will sample the rays belonging to a local region simultaneously. However, as mentioned earlier, defocus blur occurs due to the intermingling of neighboring pixels which happens due to the shallow depth of field of the camera. In order to model such interference of neighboring pixels, peripheral region of the interesting pixel must be rendered as well. Therefore, instead of conventional random ray sampling, we adopted random patch sampling that samples a group of rays within a close location, namely a patch. 
However, rendering the neighboring pixels increases the training time, and this increase is inevitable in deblurring neural fields literature. Naively reducing the number of rays to minimize the training time rather degrades the reconstruction quality. As the neural fields lack the notion of 3D consistency, we adopted the patch sampling approach to model the blurring phenomenon by learning the affinity among the neighboring pixels.

\begin{figure}[t]
\begin{center}
\includegraphics[width=1.0\linewidth]{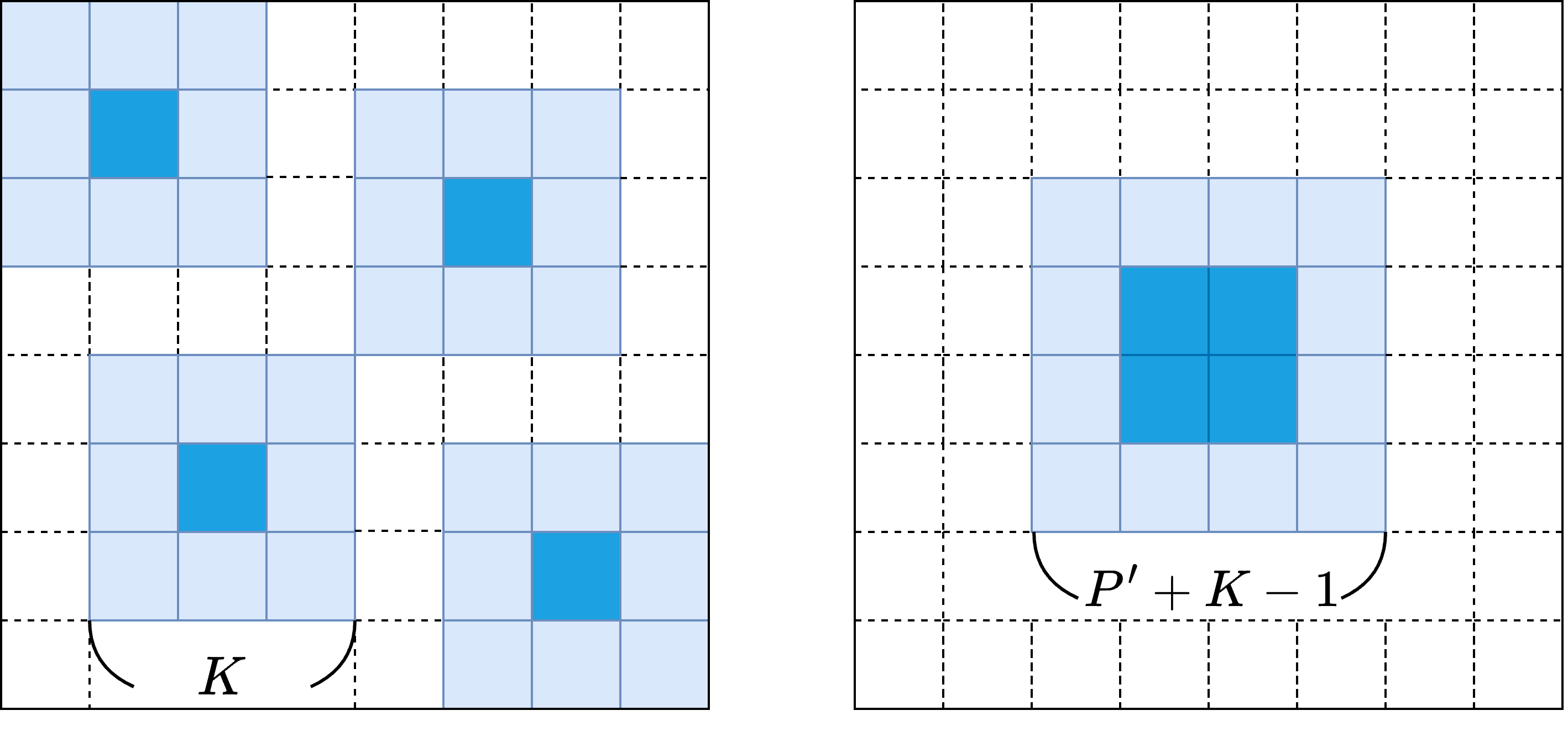}
\end{center}
\vspace{-0.5cm}
   \caption{Left: random ray sampling. Right: random patch sampling. Blue pixels are $P' \times P'$ interesting pixels to be rendered and skyblue pixels are required neighboring pixels for blur convolution.}
\label{fig:patch_sampling}
\end{figure}

\cref{fig:patch_sampling} suggests that $N \times K \times K$ pixels should be rendered to produce blurred intensity values of $N$ pixels, where $N$ is the number of target pixels and $K \times K$ is the size of blur kernel. Interestingly, under random patch sampling, target pixels can share their neighboring pixels within a patch so that the number of neighboring pixels is reduced.  Such that, only $(P' + K + 1)^2$ pixels need to be rendered to produce the same number of target pixels, where $P'$ is the width and height of region to compute intensity. The number of sampled rays is no longer dependent on $N$ but $P'^2$, so random patch sampling can effectively reduce the training time by shrinking the amount of ray rendering. This approach saves a lot of time, in contrast to NeRF that fetches thousands of rays per training iteration where $N >> P'^2$. With random patch sampling, we have reduced the number of required neighboring rays per ray from $K^2$ to  $\frac{{\left (P'+K+1 \right )}^2}{{P'}^2}$.


\subsection{Sharpness Prior} \label{SMO}     
We propose to leverage a sharpness prior to accurately model the deblurring phenomenon for sharp NeRF. By applying this sharpness prior, sharpness level of image pixels is determined. This sharpness information plays a key role in learning the suitable blur kernels for those pixels. We used the defocus map estimation network (DMENet)~\cite{Lee2019DMENet}, which is the first end-to-end CNN framework to directly estimate spatially varying defocus map from a given defocus image. We estimate defocus map of every training view and quantize it uniformly into $N_k$ values. Additionally, the sharpness prior is computed only once as part of the preprocessing, taking less than a couple of minutes, and it does not affect the subsequent training and testing times. Quantized defocus map is used as a sharpness level map $L$ which divides pixels into $N_k$ groups based on their degree of sharpness.
Here, it is worth mentioning that our proposed method is not specific to applying DMENet\cite{Lee2019DMENet}, but is generic. That is, any operator from the literature can be utilized as a sharpness prior as long as it can reliably measure the sharpness level of image pixels.

The original NeRF and many other derived works~\cite{nerf,lin2021barf,wang2021nerf} treated the image pixels and their corresponding rays independently. This approach considers that the colors/intensities of the pixels/points have association only along the ray, i.e., along the principal axis. This approach ignores any transversal relation among neighboring points or rays during volume rendering. Therefore, these works cannot mitigate the defocus effects and result in blurry rendered pixels. In contrast, we propose to exploit the transversal relation among neighboring rays as well. Specifically, we compute the sharpness level of points in their neighborhood.
Furthermore, we leverage depth information to improve the extraction of regions with sharp focus. Initially, we segment the scene based on depth information and assess the level of focus within each segment. The calculation of depth information is accomplished through the application of the MIDAS~\cite{Ranftl2021}.

\subsection{Grid-based Blur Kernel}     
We propose multiple trainable grid-based blur kernels $\mathcal{B} \in \mathbb{R}^{N_{img}\times N_k \times K \times K \times C}$  where $N_{img}$ is the number of training views, $N_k$ is the number of blur kernels, $K$ is kernel size, and $C$ is the number of channels. We will omit channel $C$ for brevity from now on.
Our grid-based blur kernels can be directly optimized to model the spatially-varying blurriness of the given scene, in contrast to the previous works~\cite{lee2022deblurred, ma2022deblur, peng2022pdrf} which employed MLP to generate blur kernels. Learnable grid-based kernels allow to obtain blur kernels $w_{x}$ of each pixel through simple indexing: 
\begin{equation}
    w_{x} = F(\mathcal{B}, L, x),
\end{equation}
where $F$ is a mapping function that maps each pixel to corresponding blur kernel $w_{x} \in \mathbb{R}^{P'\times P' \times K \times K}$ based on sharpness level $L$. Then we crop the clean image $I_c$ along to height and width with stride 1 and get cropped image $I'_c \in \mathbb{R}^{P'\times P' \times K \times K}$. Using the obtained $w_{x}$ and $I'_c$, blurred image can be convolved from clean image as following:
\begin{align}
&I_b[h,w] = \sum_{i=0}^{K}\sum_{j=0}^{K} I'_c[h,w,i,j] w_{x}[h,w,i,j],
\end{align}
where $I_b$ is a blurred image with the shape of $P' \times P'$, $[\cdot]$ is an indexing operation, and $h$, $w$ represents index along the height and width of image and blur kernels respectively. Both blur kernels $\mathcal{B}$ and neural fields are optimized via loss $\mathcal{L}_{recon}$ in \cref{eq:mse} which is an MSE loss within $I_b$ and ground truth image $I_{gt}$:
\begin{equation}
    \mathcal{L}_{recon} = ||I_b - I_{gt}||_2.
    \label{eq:mse}
\end{equation} 
Using learnable grid-based blur kernels, we can easily improve expressivity of the blur kernels by simply increasing the size of grid which does not require extra cost to produce kernels and successfully rendered clean image even from harshly defocus region. 

\begin{figure}[]
\begin{center}
\includegraphics[width=1.0\linewidth]{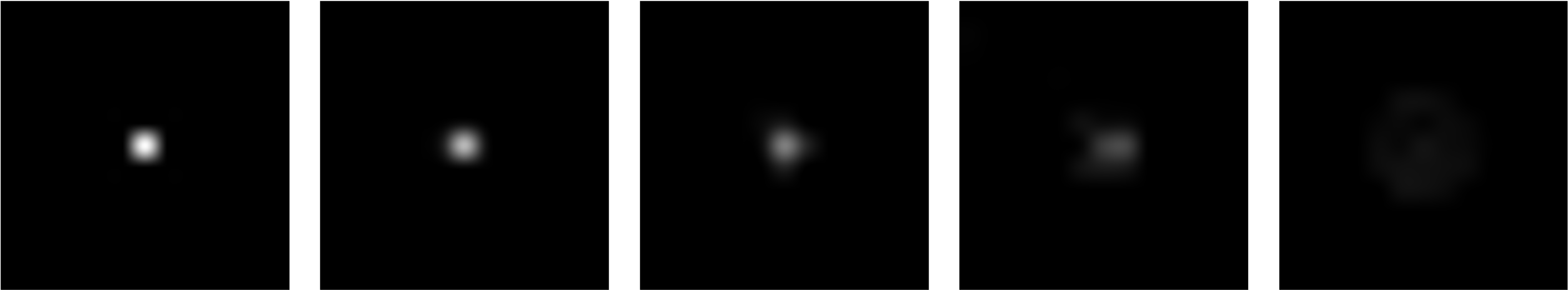}
\end{center}
\vspace{-0.5cm}
   \caption{Visualization of blur kernels. From left to right, the values of kernels are widely spread which implies that leftmost kernel is responsible for sharp region and rightmost kernel is responsible for blurry region.}
\label{fig:kernel_vis}
\end{figure}

Instead of optimizing per pixel blur kernels, we train $N_k$ blur kernels which is identical to the number of the unique groups of $L$ and pixels in a same group share a blur kernel, for effective optimization. It is worth to note that applying appropriate blur kernels to appropriate regions of the images plays pivotal role to estimate the scene blurriness because different regions in various depth of fields have varying degrees of blurriness. As pixels are blurred because of the interference of the neighboring pixels, the value of blur kernels tend to be spread widely to estimate the interruption of wide range of peripheral pixels if given pixels are severely blurry, while they tend to focus on the central part of neighboring pixels if given pixels are sharp. \cref{fig:kernel_vis} shows different form of blur kernels depending on the blurriness of their interesting regions. Therefore, each blur kernel can estimate blurriness correctly only if it is assigned to right region where the degree of blurriness is similar within pixels. Table~\ref{tab:ablation_grouping} shows the quantitative results on comparing building $L$ with sharpness prior and other approaches without sharpness prior.

\section{Experiments}
\subsection{Implementation Details}     
We used TensoRF~\cite{chen2022tensorf} as the baseline and build a grid-based learnable kernel. Grid-based blur kernels are jointly optimized with the grid-based neural fields, starting from the first training iteration, as the grid-based neural fields converge rapidly. We set the size of each blur kernel to 11, learning rate of the kernel to 0.05 and initialize with 2D Gaussian distribution. 
We use 400 for $N_k$ and skip convolution for pixels in 100 groups of each view with high sharpness value.
For random patch sampling, we sample 28 patches with the patch size $P=22$, so the number of the interesting pixels $P'^{2}$ is $12^{2}$.
We reduce the grid resolution from $640^3$ to $480^3$ and raise the number of grid component from 96 to 132.
We use the learnable camera response function proposed by~\cite{ma2022deblur}.
All the experiments were conducted on a single NVIDIA A100 GPU and please refer TensoRF~\cite{chen2022tensorf} for more details.
\subsection{Evaluation Metrics}         
Usually during the image acquisition and processing, distortions can cause an image's quality to decrease. Examples of distortion include noise, blurring, ringing, and compression artifacts. Efforts have been made to develop impartial metrics of quality. A useful quality metric corresponds favorably with the subjective assessment of quality made by a human observer in many situations.
\\\textbf{Full-reference quality metrics:}
If a distortion-free image is available, it can be used as a reference to measure the quality of other restored/reconstructed images. In such cases, full-reference quality metrics can be used to compare the target image with the reference image. We have used peak signal-to-noise ratio (PSNR) and structural similarity index (SSIM)~\cite{wang2004image}.
\\\textbf{No-reference quality metrics:}
If distortion-free reference image is not available, an alternative is to employ a no-reference image quality metric. These metrics compute the image quality based on image statistics. When compared to full-reference metrics, all no-reference quality metrics typically perform better in terms of agreement with a subjective human quality score. We have used Blind/referenceless image spatial quality evaluator (Brisque)~\cite{mittal2012no}, and Natural image quality evaluator (Niqe)~\cite{mittal2012making}. All of these quality metrics have been defined in the supplementary material.


%


\subsection{Dataset}                    
We conducted experiments on the scenes which contain real and synthetic defocus blurred images of the Deblur-NeRF dataset~\cite{ma2022deblur}. There are ten scenes in the real defocus category and five in synthetic defocus category of this dataset. They utilized the in-built skills of Blender to produce depth-of-field images in order to reduce defocus blur. Realistic defocus blur effects were obtained by establishing the aperture size and arbitrarily choosing a focal plane between the nearest and farthest depths. They select a large aperture to capture the defocus image and use COLMAP~\cite{schoenberger2016mvs, schoenberger2016sfm} to compute the camera pose of the blur and reference image in the real scene.
\subsection{Results}                    
In this section, we present full reports of experiments with both qualitative and quantitative results. 
We compare the proposed method to existing deblurring neural fields models, Deblur-NeRF~\cite{chen2022tensorf}, DP-NeRF~\cite{lee2022deblurred}, and PDRF~\cite{peng2022pdrf}. We also compare ours with TensoRF~\cite{chen2022tensorf} trained with deblurred image using Restormer~\cite{Zamir2021Restormer}. 
Restormer is a modern Transformer-based model that showed noticeable image restoration performance. 
Table~\ref{tab:res_imgref} shows the quantitative results on real defocus dataset evaluated using full-reference quality metrics PSNR and SSIM. We also evaluated our model using no-reference quality metrics, Brisque and Niqe as shown in Table~\ref{tab:res_noneimgref}. Sharp-NeRF achieves comparable results to other MLP-based models consistently under various image quality evaluation metrics and even attains state-of-the-art performance under Niqe metric while reducing training time to less than a half hour which is the shortest training time among all deblurring neural fields models.
Due to the page limit, results on synthetic data and other experiments will be covered in supplementary materials.
\begin{table*}[t]
    \centering
    \caption{Quantitative results on real defocus dataset under Full-reference metrics. Each cell is colored as \colorbox{orange!50}{best} and \colorbox{yellow!50}{second best}.}
    \resizebox{\linewidth}{!}{
    \begin{tabular}{c|cc|cc|cc|cc|cc|cc|c}
    \hline
         & \multicolumn{2}{|c|}{Cake}& \multicolumn{2}{|c|}{Caps} &  \multicolumn{2}{|c|}{Cisco} & \multicolumn{2}{|c|}{Coral} & \multicolumn{2}{|c|}{Cupcake}& \\
     & PSNR$\uparrow$ & SSIM$\uparrow$ & PSNR$\uparrow$ & SSIM$\uparrow$ & PSNR$\uparrow$ & SSIM$\uparrow$ & PSNR$\uparrow$ & SSIM$\uparrow$ & PSNR$\uparrow$ & SSIM$\uparrow$ \\
     \hline
     \hline
    NeRF~\cite{nerf} & 24.42 & 0.72 & 22.73 & 0.63 & 20.72 & 0.72 & 19.81 & 0.56 & 21.88 & 0.68 \\
    TensoRF~\cite{chen2022tensorf} & 24.08 & 0.70 & 21.55 & 0.53 & 20.56 & 0.70 & 19.27 & 0.54 & 21.16 & 0.65 \\
    Restormer~\cite{Zamir2021Restormer}+TensoRF & 23.96 & 0.68 & 21.54 & 0.51 & 20.28 & 0.69 & 19.15 & 0.52 & 21.37 & 0.65  \\
    \hline
    Deblur-NeRF~\cite{ma2022deblur} & 26.27 & 0.78 & 23.87 & 0.71 & \cellcolor{yellow!50}20.83 & \cellcolor{yellow!50}0.72 & 19.85 & 0.59 & 22.26 & 0.72 \\
    DP-NeRF~\cite{lee2022deblurred} & 26.16 & 0.77 & 23.95 & 0.71 & 20.73 & 0.72 & \cellcolor{orange!50}20.11 & \cellcolor{orange!50}0.61 & 22.80 & 0.74 \\
    PDRF-5~\cite{peng2022pdrf} & \cellcolor{yellow!50}27.03 & \cellcolor{yellow!50}0.79 & \cellcolor{orange!50}24.29 & \cellcolor{orange!50}0.72 & 20.67 & 0.72 & 19.65 & 0.58 & \cellcolor{yellow!50}22.94 & \cellcolor{orange!50}0.74 \\
    PDRF-10~\cite{peng2022pdrf} & \cellcolor{orange!50}27.06 & \cellcolor{orange!50}0.80 & \cellcolor{yellow!50}24.06 & \cellcolor{yellow!50}0.71 & 20.68 & 0.72 & 19.61 & 0.58 & \cellcolor{orange!50}22.95 & \cellcolor{yellow!50}0.74 \\
    Ours & 26.23 & 0.77 & 23.98 & 0.70 & \cellcolor{orange!50}20.88 & \cellcolor{orange!50}0.72 & \cellcolor{yellow!50}20.07 & \cellcolor{yellow!50}0.59 & 22.75 & 0.73 \\
    \hline
         & \multicolumn{2}{|c|}{Cups}& \multicolumn{2}{|c|}{Daisy} &  \multicolumn{2}{|c|}{Sausage} & \multicolumn{2}{|c|}{Seal} & \multicolumn{2}{|c}{Tools}& \multicolumn{2}{|c|}{Average} & Time \\
     & PSNR$\uparrow$ & SSIM$\uparrow$ & PSNR$\uparrow$ & SSIM$\uparrow$ & PSNR$\uparrow$ & SSIM$\uparrow$ & PSNR$\uparrow$ & SSIM$\uparrow$ & PSNR$\uparrow$ & SSIM$\uparrow$ & PSNR$\uparrow$ & SSIM$\uparrow$ & Hours $\downarrow$\\
     \hline
     \hline
    NeRF~\cite{nerf} & 25.02 & 0.75 & 22.74 & 0.62 & 17.79 & 0.48 & 22.79 & 0.62 & 26.08 & 0.85 & 22.40 & 0.66 & 2.30 \\
    TensoRF~\cite{chen2022tensorf} & 23.56& 0.70 & 22.74 & 0.63 & 17.04 & 0.44 & 21.19 & 0.55 & 21.75 & 0.74 & 21.29 & 0.62 & \cellcolor{orange!50}0.39\\
    Restormer~\cite{Zamir2021Restormer}+TensoRF & 23.03 & 0.68 & 22.53 & 0.62 & 16.59 & 0.42 & 20.52 & 0.52 & 23.78 & 0.79 & 21.28 & 0.61 & \cellcolor{orange!50}0.39 \\
    \hline
    Deblur-NeRF~\cite{ma2022deblur} & 26.21 & 0.79 & 23.52 & 0.68 & 18.01 & 0.49 & 26.04 & 0.77 & 27.81 & 0.89 & 23.46 & 0.71 & 10.35 \\
    DP-NeRF~\cite{lee2022deblurred} & \cellcolor{orange!50}26.75 & \cellcolor{orange!50}0.81 & 23.79 & 0.69 & 18.35 & 0.54 & 25.95 & 0.77 & \cellcolor{orange!50}28.07 & 0.89 & 23.67 & 0.72 & 20.20 \\
    PDRF-5~\cite{peng2022pdrf} & 26.37 & 0.80 & \cellcolor{yellow!50}24.30 & \cellcolor{yellow!50}0.73 & \cellcolor{yellow!50}18.82 & \cellcolor{yellow!50}0.55 & \cellcolor{yellow!50}26.12 & \cellcolor{yellow!50}0.78 & \cellcolor{yellow!50}28.00 & 0.89 & \cellcolor{yellow!50}23.82 & \cellcolor{yellow!50}0.73 & 1.00 \\
    PDRF-10~\cite{peng2022pdrf} & \cellcolor{yellow!50}26.39 & \cellcolor{yellow!50}0.80 & \cellcolor{orange!50}24.49 & \cellcolor{orange!50}0.74 & \cellcolor{orange!50}18.94 & \cellcolor{orange!50}0.56 & \cellcolor{orange!50}26.36 & \cellcolor{orange!50}0.79 & \cellcolor{yellow!50}28.00 & \cellcolor{yellow!50}0.89 & \cellcolor{orange!50}23.85 & \cellcolor{orange!50}0.73 & 1.67 \\
    Ours & 25.34 & 0.77 & 23.66 & 0.70 & 18.77 & 0.54 & 25.82 & 0.77 & 27.98 & \cellcolor{orange!50}0.90 & 23.55 & 0.72 & \cellcolor{yellow!50}0.48 \\
    \hline
    \end{tabular}
    }
    \label{tab:res_imgref}
\end{table*}

\begin{table*}[t]
    \centering
    \caption{Quantitative results on real defocus dataset under No-reference metrics. Each cell is colored as \colorbox{orange!50}{best} and \colorbox{yellow!50}{second best}.}
    \resizebox{\linewidth}{!}{
    \begin{tabular}{c|cc|cc|cc|cc|cc|cc}
    \hline
         & \multicolumn{2}{|c|}{Cake}& \multicolumn{2}{|c|}{Caps} &  \multicolumn{2}{|c|}{Cisco} & \multicolumn{2}{|c|}{Coral} & \multicolumn{2}{|c|}{Cupcake}&\\
     & Brisque$\downarrow$ & Niqe$\downarrow$ & Brisque$\downarrow$ & Niqe$\downarrow$ & Brisque$\downarrow$ & Niqe$\downarrow$ & Brisque$\downarrow$ & Niqe$\downarrow$ & Brisque$\downarrow$ & Niqe$\downarrow$ \\
     \hline
     \hline
    TensoRF~\cite{chen2022tensorf} & 30.27 & \cellcolor{orange!50}3.11 & 33.90 & \cellcolor{orange!50}3.08 & 35.03 & 3.83 & \cellcolor{yellow!50}36.15 & \cellcolor{orange!50}3.45 & 38.24 & 3.82 \\
    Deblur-NeRF~\cite{ma2022deblur} & 25.78 & \cellcolor{yellow!50}3.23 & \cellcolor{yellow!50}28.8 & 3.33 & 31.89 & 3.86 & 39.18 & 4.43 & 33.68 & \cellcolor{yellow!50}3.51 \\
    DP-NeRF~\cite{lee2022deblurred} & \cellcolor{yellow!50}24.15 & 3.24 & \cellcolor{orange!50}26.52 & 3.34 & \cellcolor{yellow!50}30.22 & \cellcolor{yellow!50}3.68 & 36.77 & 4.37 & 32.43 & 3.46\\
    PDRF-5~\cite{peng2022pdrf} & 25.00 & 3.60 & 37.37 & 3.58 & 30.66 & 3.97 & 35.75 & 4.13 & 32.17 & 3.71 \\
    PDRF-10~\cite{peng2022pdrf} & \cellcolor{orange!50}21.93 & 3.50 & 30.46 & 3.40 & 30.39 & 3.84 & 38.07 & 3.99 & \cellcolor{yellow!50}31.52 & 3.59 \\
    Ours & 27.54 & 3.26 & 31.66 & \cellcolor{yellow!50}3.22 & \cellcolor{orange!50}28.56 & \cellcolor{orange!50}3.63 & \cellcolor{orange!50}29.43\cellcolor{orange!50} & \cellcolor{yellow!50}3.59 & \cellcolor{orange!50}31.05 & \cellcolor{orange!50}3.49 \\
    \hline
         & \multicolumn{2}{|c|}{Cups}& \multicolumn{2}{|c|}{Daisy} &  \multicolumn{2}{|c|}{Sausage} & \multicolumn{2}{|c|}{Seal} & \multicolumn{2}{|c}{Tools}& \multicolumn{2}{|c}{Average}\\
     & Brisque$\downarrow$ & Niqe$\downarrow$ & Brisque$\downarrow$ & Niqe$\downarrow$ & Brisque$\downarrow$ & Niqe$\downarrow$ & Brisque$\downarrow$ & Niqe$\downarrow$ & Brisque$\downarrow$ & Niqe$\downarrow$ & Brisque$\downarrow$ & Niqe$\downarrow$ \\
     \hline
     \hline
    TensoRF~\cite{chen2022tensorf} & 37.41 & 3.58 & 36.79 & \cellcolor{yellow!50}4.07 & 36.86 & 3.56 & 34.72 & \cellcolor{orange!50}3.18 & 39.65 & 3.49 & 35.90 & \cellcolor{yellow!50}3.52 \\
    Deblur-NeRF~\cite{ma2022deblur} & 32.65 & 3.40 & 29.10 & 4.11 & 28.63 & \cellcolor{orange!50}3.37 & 32.09 & 3.46 & \cellcolor{yellow!50}31.44 & 3.14 & 31.32 & 3.58 \\
    DP-NeRF~\cite{lee2022deblurred} & \cellcolor{yellow!50}31.00 & \cellcolor{yellow!50}3.33 & 30.01 & 4.09 & 30.17 & \cellcolor{yellow!50}3.38 & \cellcolor{yellow!50}31.82 & 3.29 & \cellcolor{orange!50}29.86 & \cellcolor{orange!50}3.05 & 30.29 & \cellcolor{yellow!50}3.52\\
    PDRF-5~\cite{peng2022pdrf} & 33.75 & 3.60 & 29.18 & 4.37 & 29.38 & 3.60 & 32.78 & 3.45 & 35.73 & 3.44 & 32.18 & 3.75 \\
    PDRF-10~\cite{peng2022pdrf} & \cellcolor{orange!50}29.09 & 3.46 & \cellcolor{yellow!50}28.40 & 4.24 & \cellcolor{orange!50}26.67 & 3.49 & \cellcolor{orange!50}29.31 & 3.50 & 34.28 & 3.29 & \cellcolor{orange!50}30.01 & 3.63 \\
    Ours & 33.70 & \cellcolor{orange!50}3.29 & 26.35 \cellcolor{orange!50}& \cellcolor{orange!50}3.85 & \cellcolor{yellow!50}28.51 & 3.45 & 32.63 & \cellcolor{yellow!50}3.19 & 33.10 & \cellcolor{yellow!50}3.13 & \cellcolor{yellow!50}30.25 & \cellcolor{orange!50}3.41\\
    \hline
    \end{tabular}}
    \label{tab:res_noneimgref}
\end{table*}

\begin{figure*}[t]
\begin{center}
\includegraphics[width=1.0\linewidth]{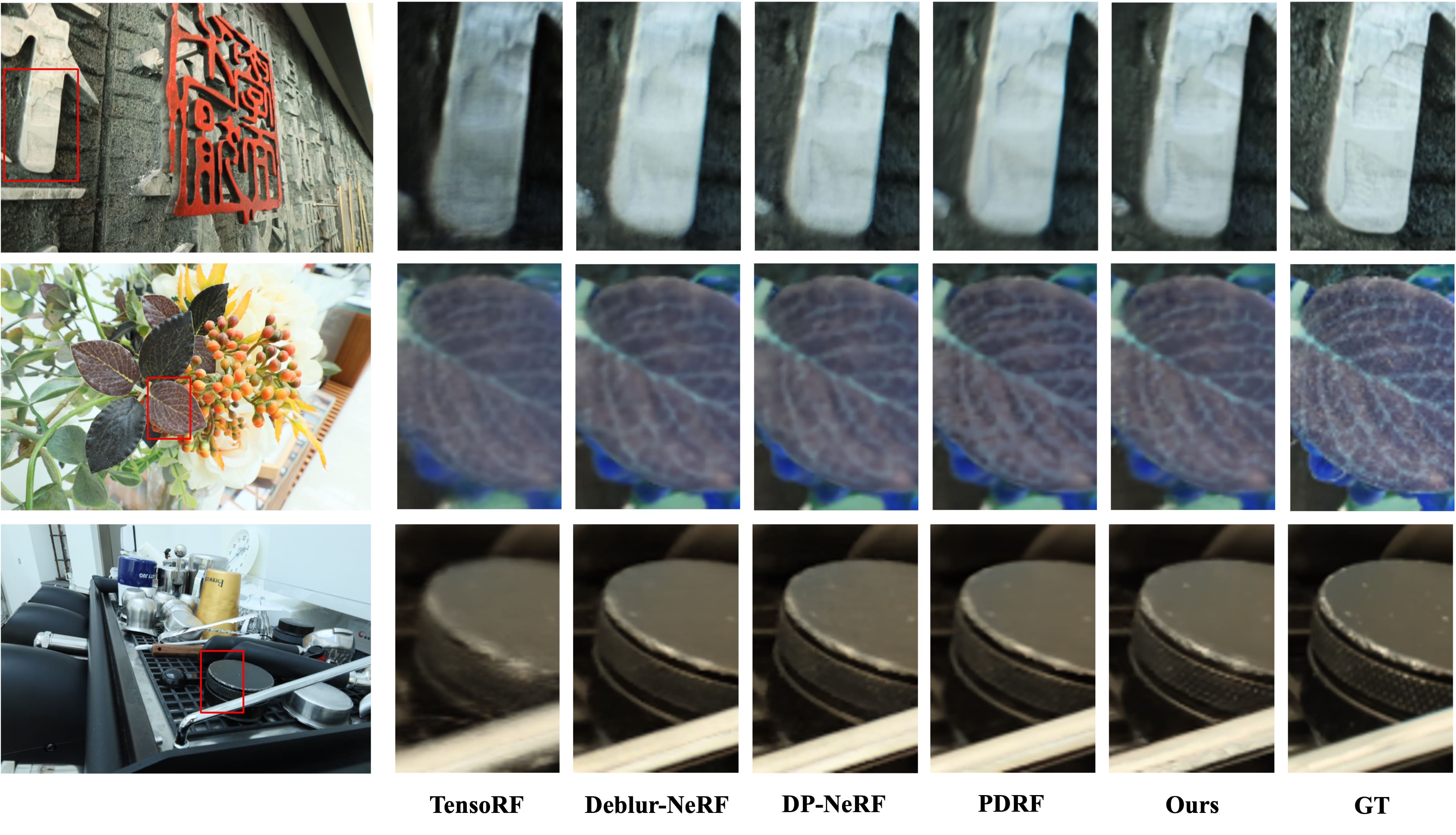}
\end{center}
\vspace{-0.5cm}
   \caption{Qualitative results on real defocus dataset. Our proposed method renders sharp images which have vivid colors and fine details.}
\label{fig:qualitative_result}
\end{figure*}

\subsection{Ablation Studies}

\noindent\textbf{Sharpness Prior.} Firstly, we conducted an ablation study on the selection of focus measure operator. We compared not only the hand-crafted operators like sum modified laplacian (SML)~\cite{nayar1994shape} and Tenengrad~\cite{krotkov1988focusing}, but also deep neural network-based model DMENet~\cite{Lee2019DMENet}. The results are presented in Table~\ref{tab:ablation_grouping}. The better values of PSNR and SSIM for DMENet~\cite{Lee2019DMENet} indicate that this sharpness prior is more effective in accurately measuring the sharpness levels of image pixels.

\noindent\textbf{Sharpness Level.} Next, we conducted experiments on building sharpness level map $L$. It can be observed in Table~\ref{tab:ablation_sharpness_level} that constructing $L$ using sharpness measurement is a major factor that considerably affects the rendering quality.

\begin{table}[h!]
    \centering
    \caption{Ablation study on selection of the sharpness prior. Average values for the real defocus testset have been mentioned.}
    \begin{tabular}{c|c|c}
    \hline
    Methods & PSNR $\uparrow$ & SSIM $\uparrow$ \\
    \hline\hline
    SML~\cite{nayar1994shape} & 22.56 & 0.67 \\
    Tenengrad~\cite{krotkov1988focusing} & 22.49 & 0.66 \\
    DMENet~\cite{Lee2019DMENet} & \textbf{23.55} & \textbf{0.72} \\
    \hline
    \end{tabular}
    \label{tab:ablation_grouping}
\end{table}

\begin{table}[h!]
    \centering
    \caption{Ablation study on constructing sharpness level map. Randomly assigning (Random) and allocating pixels into the groups based on their position (Local) show degraded performance.}
    \begin{tabular}{c|c|c}
    \hline
    Methods & PSNR $\uparrow$ & SSIM $\uparrow$ \\
    \hline\hline
    Random & 22.09 & 0.66 \\
    Local & 21.43 & 0.63 \\
    Ours & \textbf{23.55} & \textbf{0.72} \\
    \hline
    \end{tabular}
    \label{tab:ablation_sharpness_level}
\end{table}

\begin{table}[h!]
    \centering
    \caption{Comparison on kernel generation time.}
    \begin{tabular}{c|c}
    \hline
    Methods & Time $\downarrow$ \\
    \hline\hline
    MLP-based kernels & 2.20 ms \\
    Ours & \textbf{0.18 ms} \\
    \hline
    \end{tabular}
    \label{tab:ablation_time}
\end{table}

\noindent \textbf{Kernel Generation Time.} Finally, we investigated the time taken by the MLP-based and grid-based methods in generating the blur kernels. To do so, we computed the time taken by the recent MLP-based blur kernels of PDRF~\cite{peng2022pdrf} and by our method. These times are computed for real defocus dataset, and their average values are mentioned in Table~\ref{tab:ablation_time}. The considerably reduced time of our method indicates its superiority in terms of computational efficiency.

\section{Limitations \& Future Works}
Although our approach, involving learnable grid-based blur kernels, has produced appreciably good quality results, yet it has limitation. For example, our method produces degraded results for motion blurred images. This is because our approach relies on the sharpness measurement of the input images, and we have incorporated a sharpness prior only for the defocus blur, and not for the motion blur. In other words, when the input images have motion blur, our sharpness prior fails in accurately estimating the sharpness of the pixels, which ultimately leads to the poor quality of rendered images. However, it is worth mentioning that this limitation is due to the design methodology of our approach. For instance, instead of defocus blur, our approach can be adjusted and applied to motion blurred scenes by incorporating a sharpness prior designed for the motion blurred images. 
We anticipate that estimation of the camera motion blur can be achieved by computing the orientation and magnitude of blurriness in the pixels using higher order derivatives, and we will explore it in our future work. Further, it seems interesting that whether few other priors, or group of priors, can be incorporated in our method to produce good quality results for other image degrading situations, like haze, raining, and under water imaging. In future, we will learn the blur kernels in 3D and we will cover all types of blur. Further, we will seek any gain that can be achieved by learning the kernels through joint optimization with the rest of the network.

\section{Conclusion}
This paper proposes first fully grid-based deblurring neural fields---Sharp-NeRF. In the literature, it has been observed that grid-based neural fields suffer from quality degradation when rendering under imperfect conditions such as defocused images~\cite{peng2022pdrf}. To overcome this shortcoming, in this work, we proposed novel learnable grid-based blur kernels. This approach can produce blur kernels with simple indexing and sharpness measurement to promote blur kernel optimization. Coupled with grid-based kernel and random patch sampling, we could achieve state-of-the-art training time, taking only a half an hour to train.
\section{Acknowledgements}
This work was supported in parts by the Institute of Information and Communication Technology Planning Evaluation (IITP) grant (IITP-2019-0-00421), Creative Challenge Research Program (2021R1I1A1A01052521) through the National Research Foundation (NRF) of Korea, and the NRF grant (RS-2023-00245342) funded by the Ministry of Science and ICT (MSIT) of Korea.


{\small
\bibliographystyle{ieee_fullname}
\bibliography{SharpNeRF}
}

\clearpage
\appendix
\renewcommand\thesection{\Alph{section}}
\setcounter{section}{0}
 
\twocolumn[
\begin{center}
\Large{\bf{Sharp-NeRF: Grid-based Fast Deblurring Neural Radiance Field using Sharpness Prior \\ Supplementary Material}}\par\vspace{3ex}
\end{center}]

\maketitle

\begin{figure*}[t!]
\begin{center}
\includegraphics[width=1.0\linewidth]{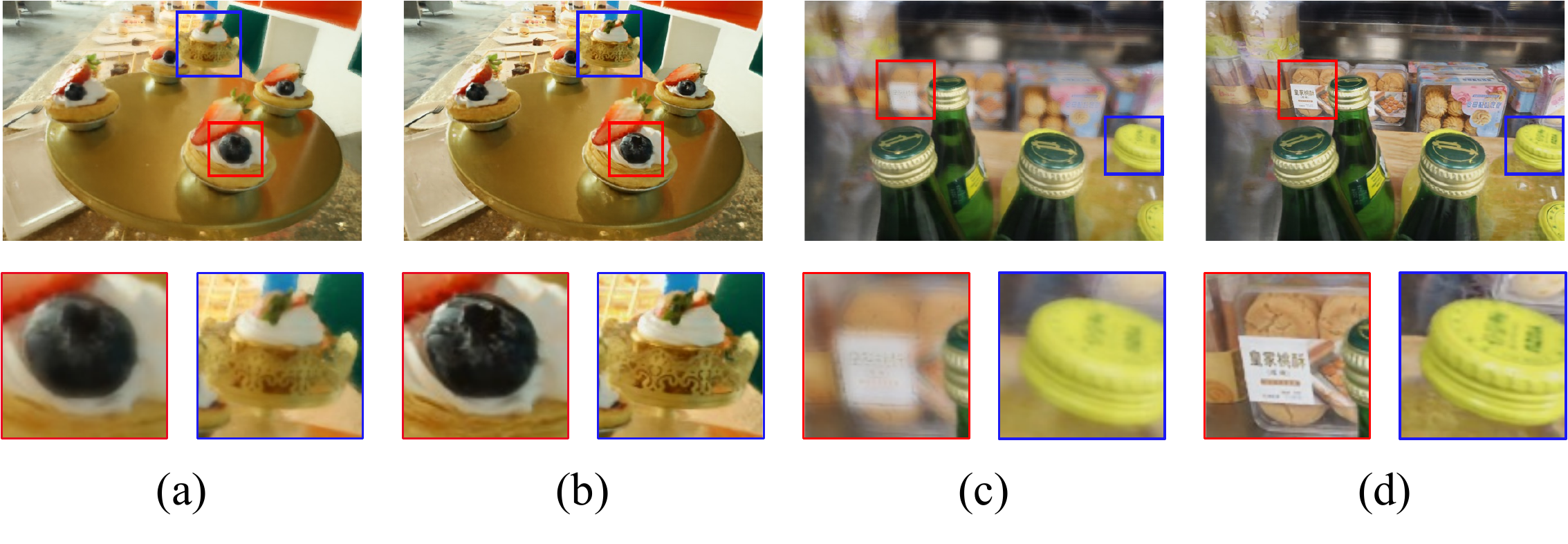}
\end{center}
\vspace{-0.5cm}
   \caption{Comparison on learanble blur kernel and fixed blur kernel. (a), (c): Our rendering module trained with fixed discrete kernels. (b), (d): Ours (with learnable discrete kernels).}
\label{fig:argmin}
\end{figure*}

In this supplementary material, we provide more details on different modules of our method and describe results of experiments that we mentioned in the main paper.
\section{Additional Ablation Studies}
\subsection{Discrete Kernel: Fixed Vs Learnable}
The recently proposed Hybrid Neural Rendering Model~\cite{dai2023hybrid} also employs several grid kernels to render clean image from the camera motion blurred images, but they are fixed. It applies convolution to clean rendered image with all blur kernels and picks the one that produces the minimal loss with the ground truth image. 
\cref{fig:argmin} shows rendered images from our method and our model that uses fixed discrete kernels as Hybrid Neural Rendering model does.
We can see that the blur kernels cannot estimate spatially varying defocus blur accurately and fail to render clean image. We assume this is because the point spread function of defocus blur is more complex than that of camera motion blur. Defocus blur depends on various tangled optical factors, and cannot be easily estimated with fixed blur kernels.

\subsection{Patch Size Analysis}
Patch size is a key factor that determines the rendering quality and the training time. Larger patches mean more shared neighbors, so they can effectively decrease training time. However, rendering quality degrades as the number of patches will decrease unless we do not raise the batch size. Table~\ref{tab:ablation_patchsize} shows the quality metrics for rendered images, the number of required rays for each pixel, and the training times required for various patch sizes $P$. The number of the target pixels $P'^2$ for $P=26$, $P=22$, $P=18$ is $16^2$, $12^2$, $8^2$, respectively. These results show that increasing the patch size reduces the number of required neighboring pixels, which in turn reduces the training time, but results in quality degradation.

\begin{table}[h]
    \centering
    \caption{Ablation study: Patch size evaluated for real defocus dataset. The average number of required neighboring rays per pixel is denoted as \#Rays.}

    \resizebox{\columnwidth}{!}{
    \begin{tabular}{c|c|c|c|c|c|c}
    \hline
    Patch Size & PSNR$\uparrow$ & SSIM$\uparrow$ & Brisque$\downarrow$ & Niqe$\downarrow$ & \#Rays & Time \\
    \hline\hline
    $P = 26$ & 23.44  & 0.7228 & \textbf{29.50} & 3.44 & \textbf{2.64} & \textbf{0.40} \\
    $P = 22$ & 23.55  & 0.7258 & 30.25 & 3.41 & 3.36 & 0.48\\
    $P = 18$ & \textbf{23.57} & \textbf{0.7501} & 29.64 & \textbf{3.38} & 5.06  & 0.57\\
    \hline
    \end{tabular}
    }

    \label{tab:ablation_patchsize}
\end{table}

\begin{table}[t]
    \centering
    \caption{Compassion of our method with PDRF with its different number of neighboring rays. Models are evaluated on real defocus dataset. `$n$' in PDRF-$n$ stands for the number of required neighboring rays. * means that we modified PDRF to analyze its performance for this setting.}

    \resizebox{\columnwidth}{!}{
    \begin{tabular}{c|c|c|c|c|c}
    \hline
    Methods & PSNR$\uparrow$ & SSIM$\uparrow$ & Brisque$\downarrow$ & Niqe$\downarrow$ & Time \\
    \hline\hline
    PDRF-5 & \textbf{23.82} & \textbf{0.7382} & 32.18 & 3.75 & 1.00 \\
    PDRF-3* & 23.50 & 0.7217 & 33.35 & 3.92 & 0.88 \\
    PDRF-2* & 22.34 & 0.6713 & 36.74 & 4.09 & 0.70 \\
    Ours & 23.55 & 0.7258 & \textbf{30.25} & \textbf{3.41} & \textbf{0.48} \\
    \hline
    \end{tabular}
    }

    \label{tab:ablation_pdrf}
\end{table}

\subsection{Comparison with Variants of PDRF~\cite{peng2022pdrf}}
In this subsection, we compare our proposed method Sharp-NeRF with the very recently proposed method PDRF~\cite{peng2022pdrf}. We compared them in terms of their training time and the image quality of their rendered images using various metrics. As the required number of neighboring rays is set to 5 in PDRF~\cite{peng2022pdrf}, PDRF-5 denotes this actual work. As the required number of neighboring rays is a key factor for training time, we reduced the number of neighboring rays of PDRF to decrease its training time to as low as ours. Table~\ref{tab:ablation_pdrf} implies that our method requires significantly lesser training time, and importantly, without any noticeable rendering quality degradation compared to PDRF.

\begin{figure*}[t]
\begin{center}
\includegraphics[width=1.0\linewidth]{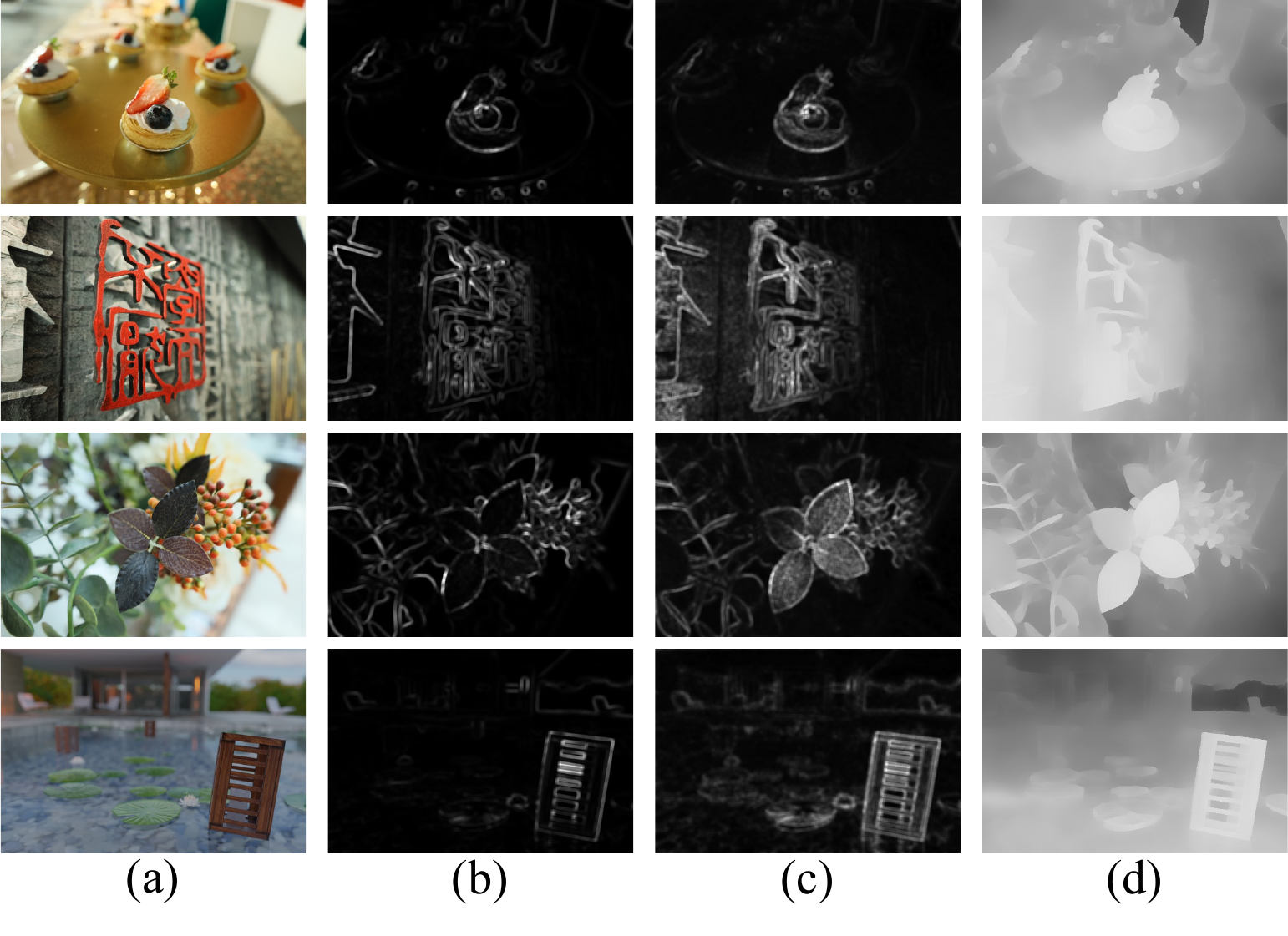}
\end{center}
   \caption{Comparison of sharpness maps provided by various sharpness priors. (a): Sample images, and sharpness maps from (b): Tenengrad, (c): SML, and (d): DMENet.}
\label{fig:focus_map}
\end{figure*}

\begin{figure*}[t!]
\begin{center}
\includegraphics[width=1.0\linewidth]{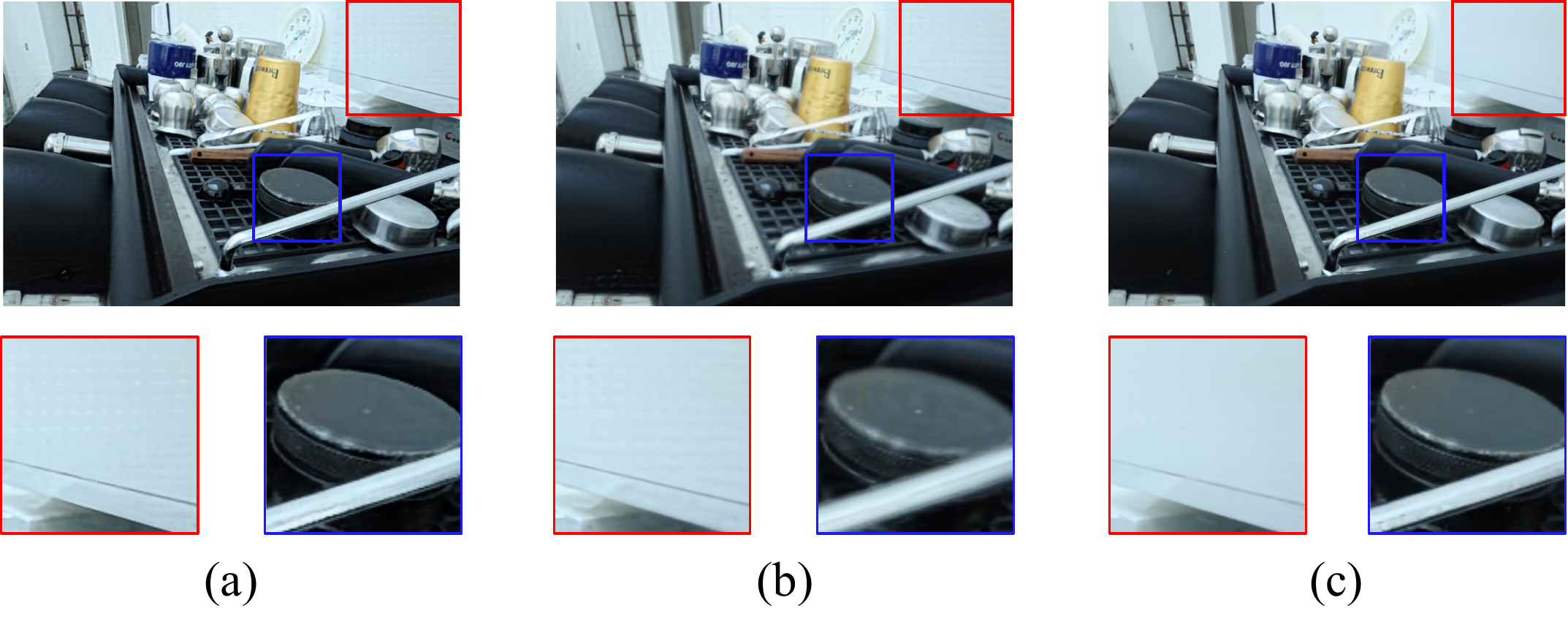}
\end{center}
\vspace{-0.5cm}
   \caption{Comparison of various sharpness priors in terms of the quality of rendered images. (a): SML, (b): Tenengrad, (c): DMENet.}
\label{fig:fmo}
\end{figure*}

\section {Selection of Sharpness Prior}
In our proposed method Sharp-NeRF, sharpness prior has been used to measure the sharpness level of input image pixels. These sharpness measurements determine how much effort will be put in learning the discrete kernels for those pixels. 
We have tested three sharpness priors in our study. Among these priors, two are very famous focus measure operators. These operators are sum modified Laplacian (SML)~\cite{nayar1994shape} and Tenenbaum gradient (Tenengrad)~\cite{krotkov1988focusing}, and these are hand-crafted. The third sharpness prior is DMENet~\cite{Lee2019DMENet} which is deep learning-based and it has been proposed recently.
 
\cref{fig:focus_map} displays sharpness maps of different sharpness priors. Depending on their working principle, SML and Tenengrad measure sharpness using derivatives of pixels which becomes hard to compute on textureless regions. This is indicated by the blue and red boxes, respectively. It's evident that SML and Tenengrad tend to focus on sharp regions primarily along edges. In contrast, DMENet does a better work in distinguishing between sharp and blurry regions even there is no or little texture. On the basis of this observation, we decided to use DMENet~\cite{Lee2019DMENet} as our sharpness prior. \cref{fig:fmo} shows a qualitative result for the performance of these three priors.
 

\begin{table*}[t]
\centering
\caption{Full-reference and No-reference image quality metrics.}
\label{tab:QM_definations}
\begin{tabular}{@{}ll@{}}
\toprule
PSNR $\uparrow$  &  Peak signal-to-noise ratio. It is derived from MSE and indicates the ratio of the \\ &  maximum pixel intensity to the power of the distortion. \\
SSIM~\cite{wang2004image} $\uparrow$  &  Structural similarity index. It combines the local image structure, brightness, and \\ &  contrast into a single local quality score \\ \midrule
Brisque~\cite{mittal2012no} $\downarrow$  &  Blind/referenceless image spatial quality evaluator. This model is trained on a database \\ &  of images with known distortions and as a result it can only assess the quality of images \\ &  with the same kind of distortion.  \\
Niqe~\cite{mittal2012making} $\downarrow$  &  Natural image quality evaluator. Despite being trained on a database of reference \\ &  images, this model can assess the quality of images that have arbitrary distortion. \\
\bottomrule
\end{tabular}
\end{table*}

\section{Results on Synthetic Data}
Table~\ref{tab:res_syn_imgref} and~\ref{tab:res_syn_noimgref} show the quantitative results of experiments conducted on synthetic defocus dataset. We evaluated both Full-reference (PSNR, SSIM) and No-reference metrics (Brisque, Niqe). Our rendering quality under Full-reference metrics is not as good as compared to the result on real deblur dataset. However, our method achieved sound performance on par with state-of-the-art methods under no-refernce metrics. On top of that, this all good performance has been achieved while keeping the training time to less than half an hour, which is the shortest time among all deblurring neural fields. In short, our method has set a new state-of-the-art time for the deblurring neural fields.
\begin{table*}[t]
    \centering
    \caption{Quantitative results on synthetic defocus dataset under Full-reference metrics. Each cell is colored as \colorbox{orange!50}{best} and \colorbox{yellow!50}{second best}.}

    \resizebox{\linewidth}{!}{
    \begin{tabular}{c|cc|cc|cc|cc|cc|cc|cc|c}
    \hline
         & \multicolumn{2}{|c|}{Cozy2room}& \multicolumn{2}{|c|}{Factory} &  \multicolumn{2}{|c|}{Pools} & \multicolumn{2}{|c|}{Tanabata} & \multicolumn{2}{|c|}{Trolley}&  \multicolumn{2}{|c|}{Average} &Time \\
     & PSNR$\uparrow$ & SSIM$\uparrow$ & PSNR$\uparrow$ & SSIM$\uparrow$ & PSNR$\uparrow$ & SSIM$\uparrow$ & PSNR$\uparrow$ & SSIM$\uparrow$ & PSNR$\uparrow$ & SSIM$\uparrow$ & PSNR$\uparrow$ & SSIM$\uparrow$ &Hours \\
     \hline
     \hline
    NeRF~\cite{nerf} & 30.03 & 0.8926 & 25.36 & 0.7847 & 27.77 & 0.7266 & 23.90 & 0.7811 & 22.67 & 0.7103 & 25.93 & 0.7791 & 2.30 \\
    TensoRF~\cite{chen2022tensorf} & 30.04 & 0.8949 & 25.27 & 0.7966 & 26.81 & 0.6658 & 22.73 & 0.7701 & 22.18 & 0.7143 & 25.40 & 0.7683 & \colorbox{orange!50}{0.42} \\
    \hline
    Deblur-NeRF~\cite{ma2022deblur} & 31.85 & 0.9175 & 28.03 & 0.8628 & 30.52 & 0.8246 & 26.26 & 0.8517 & 25.18 & 0.8067 & 28.37 & 0.8527 & 10.35 \\
    DP-NeRF~\cite{lee2022deblurred} & \colorbox{yellow!50}{32.11} & 0.9215 & 29.26 & 0.8793 & \colorbox{orange!50}{31.44} & \colorbox{orange!50}{0.8529} & 27.05 & 0.8635 & 26.79 & 0.8395 & 29.33 & 0.8713 & 20.20 \\
    PDRF-5~\cite{peng2022pdrf} & 32.10 & \colorbox{yellow!50}{0.9269} & \colorbox{yellow!50}{30.34} & \colorbox{yellow!50}{0.9032} & 30.48 & 0.8262 & \colorbox{yellow!50}{27.31} & \colorbox{yellow!50}{0.8818} & \colorbox{yellow!50}{27.05} & \colorbox{yellow!50}{0.8581} & \colorbox{yellow!50}{29.46} & \colorbox{yellow!50}{0.8792} & 1.00 \\
    PDRF-10~\cite{peng2022pdrf} & \colorbox{orange!50}{32.29} & \colorbox{orange!50}{0.9305} & \colorbox{orange!50}{30.90} & \colorbox{orange!50}{0.9138} & \colorbox{yellow!50}{30.97} & \colorbox{yellow!50}{0.8408} & \colorbox{orange!50}{28.18} & \colorbox{orange!50}{0.9006} & \colorbox{orange!50}{28.07} & \colorbox{orange!50}{0.8799} & \colorbox{orange!50}{30.08} & \colorbox{orange!50}{0.8931} & 1.67 \\
    Ours & 31.32 & 0.9133 & 28.67 & 0.8979 & 30.51 & 0.8264 & 24.95 & 0.8536 & 26.03 & 0.8498 & 28.30 & 0.8682 & \colorbox{yellow!50}{0.43} \\
    \hline
    \end{tabular}
    }

\label{tab:res_syn_imgref}
\end{table*}

\begin{table*}[t!]
    \centering
    \caption{Quantitative results on synthetic defocus dataset under No-reference metrics. Each cell is colored as \colorbox{orange!50}{best} and \colorbox{yellow!50}{second best}.}

    \resizebox{\linewidth}{!}{
    \begin{tabular}{c|cc|cc|cc|cc|cc|cc|cc}
    \hline
         & \multicolumn{2}{|c|}{Cozy2room}& \multicolumn{2}{|c|}{Factory} &  \multicolumn{2}{|c|}{Pools} & \multicolumn{2}{|c|}{Tanabata} & \multicolumn{2}{|c|}{Trolley}& \multicolumn{2}{|c|}{Average}\\
     & Brisque$\downarrow$ & Niqe$\downarrow$ & Brisque$\downarrow$ & Niqe$\downarrow$ & Brisque$\downarrow$ & Niqe$\downarrow$ & Brisque$\downarrow$ & Niqe$\downarrow$ & Brisque$\downarrow$ & Niqe$\downarrow$& Brisque$\downarrow$ & Niqe$\downarrow$\\
     \hline
     \hline
    TensoRF~\cite{chen2022tensorf} & 27.52 & 3.13 & 42.26 & 3.77 & 44.27 & 4.29 & 42.29 & 3.65 & 36.56 & 3.65 & 38.58 & 3.70 \\
    \hline
    Deblur-NeRF~\cite{ma2022deblur} & \colorbox{yellow!50}{17.85} & \colorbox{orange!50}{3.09} & 35.60 & 3.57 & 39.93 & 4.08 & 37.20 & 3.63 & 35.35 & 3.56 & 33.19 & 3.59 \\
    DP-NeRF~\cite{lee2022deblurred} & \colorbox{orange!50}{17.83} & 3.20 & 35.21 & \colorbox{yellow!50}{3.54} & \colorbox{yellow!50}{38.33} & \colorbox{yellow!50}{3.89} & 36.37 & \colorbox{orange!50}{3.41} & \colorbox{orange!50}{33.95} & \colorbox{yellow!50}{3.39} & \colorbox{yellow!50}{32.34} & \colorbox{orange!50}{3.49} \\
    PDRF-5~\cite{peng2022pdrf} & 21.19 & \colorbox{yellow!50}{3.11} & \colorbox{orange!50}{27.96} & 3.74 & 44.27 & 4.62 & 36.75 & 4.00 & 34.94 & 3.46 & 33.02 & 3.79\\
    PDRF-10~\cite{peng2022pdrf} & 19.65 & 3.17 & \colorbox{yellow!50}{28.24} & 3.58 & 40.42 & 4.37 & \colorbox{orange!50}{35.97} & 3.81 & \colorbox{yellow!50}{34.40} & \colorbox{orange!50}{3.37} & \colorbox{orange!50}{31.74} & 3.66\\
    Ours & 21.65 & 3.22 & 35.19 & \colorbox{orange!50}{3.44} & \colorbox{orange!50}{36.91} & \colorbox{orange!50}{3.87} & \colorbox{yellow!50}{36.10} & \colorbox{yellow!50}{3.58} & 34.77 & 3.44 & 32.92 & \colorbox{yellow!50}{3.51} \\
    \hline
    \end{tabular}
    }

    \label{tab:res_syn_noimgref}
\end{table*}


\section{No-Reference Metrics for Deblurring NeRF}
The commercially available optical imaging devices (cameras) have limited depth of field. Using these cameras, usually, when an image is captured of a scene that has a larger depth of field, some parts of the scene look blurred in the image.
Unless, the camera is tuned for the appropriate larger depth of field, or some possessing like image stitching or defocus deconvolution is performed, those regions of the scene that fall out of the range of depth of field of the camera, appear blurred in the images. 
In contrast, an image in which all the regions of the captured scene are sharp is called an all-in-focus (AIF) image. Although, it is possible to acquire defocused and all-in-focus pair by DSLR camera in two sequential shots, such as~\cite{abuolaim2020defocus} which collects 500 defocused and all-in-focus pairs by dual-pixel (DP) camera via adjusting aperture size and exposure time in two separate shots. However, datasets collected by such methods suffer from several artifacts such as inconsistent brightness and mismatched contents. In other words, obtaining a reliable AIF and defocus image pair is very challenging. This is the reason why, until recently, there has been a lack of real image datasets that contain defocused and AIF image pairs~\cite{ruan2021aifnet}. The same difficulty arises for NeRF frameworks. The dataset that has been used in our study is the Deblur-NeRF dataset~\cite{ma2022deblur}. In this dataset, the defocus blurred images were also captured using the large aperture, and hence they are not free from errors. Owing to the mismatch between AIF and defocus blurred images, deblurring techniques may suffer from inconsistent results when employing the full-reference metrics. The inconsistency among full-reference metrics (especially PSNR) has been observed and highlighted in the recent deblurring NeRFs~\cite{dai2023hybrid,lin2023vision}. For example, it has been mentioned in \cite{dai2023hybrid} that if the reference images are blurry, the PSNR (and SSIM) values may be worse. 
It has been observed in \cite{lin2023vision} that few blurry images can still achieve higher PSNR than their sharp (deblurred) counterparts. Due to these reasons, we have opted to include the No-reference metrics as well. When compared to Full-reference metrics, generally, all No-reference quality metrics typically perform better in terms of agreement with a subjective human quality score. 

\end{document}